\pdfoutput=1
\documentclass[leqno,onefignum,onetabnum]{siamltex}
\usepackage[utf8]{inputenc} % allow utf-8 input
\usepackage[T1]{fontenc}    % use 8-bit T1 fonts
\usepackage{url}            % simple URL typesetting
\usepackage{booktabs}       % professional-quality tables
\usepackage{amsfonts}       % blackboard math symbols
\usepackage{nicefrac}       % compact symbols for 1/2, etc.
\usepackage{microtype}      % microtypography
\usepackage{soul}
\usepackage[utf8]{inputenc}
\usepackage{graphicx}
\usepackage{amsmath}
\usepackage{booktabs}

\usepackage[left=2.5cm,right=2.5cm, top=2cm, bottom=2cm]{geometry}

\usepackage[colorlinks=true,
citecolor=blue,
filecolor=black,
linkcolor=blue,
urlcolor=blue]{hyperref}

\usepackage{microtype}
\usepackage{graphicx}
\usepackage{subfigure}
\usepackage{booktabs} % for professional tables

\usepackage{amsmath,amsfonts,bm}

\usepackage{xcolor}
\usepackage{colortbl}

\def\secref#1{section~\ref{#1}}
\def\eqref#1{equation~\ref{#1}}
\def\1{\bm{1}}

\DeclareMathAlphabet{\mathsfit}{\encodingdefault}{\sfdefault}{m}{sl}
\SetMathAlphabet{\mathsfit}{bold}{\encodingdefault}{\sfdefault}{bx}{n}

\def\gJ{{\mathcal{J}}}

\def\gL{{\mathcal{L}}}

\usepackage{multirow}
\usepackage{paralist}

\usepackage{booktabs} % for professional tables
\usepackage{multicol}
\usepackage{diagbox}

\usepackage[ruled,noend]{algorithm2e}

\SetCommentSty{mycommfont}

\usepackage{here}

\usepackage{amsmath,amssymb,amsfonts,amsbsy,amsfonts,latexsym}
\usepackage{multirow}
\usepackage{makecell}
\usepackage[labelfont=bf,textfont=it,belowskip=0pt,aboveskip=5pt,tableposition=top]{caption}
\usepackage{xcolor}
\usepackage{colortbl}

\definecolor{colorA}{RGB}{189,201,225}
\definecolor{colorB}{RGB}{103,169,207}
\definecolor{colorC}{RGB}{ 28,144,153}
\definecolor{colorD}{RGB}{  1,108, 89}

\newcolumntype{R}{>{\columncolor{gray!40}}r}
\newcolumntype{L}{>{\columncolor{gray!40}}l}
\newcolumntype{C}{>{\columncolor{gray!40}}c}

\usepackage{tabularx,colortbl,xcolor}
\usepackage{multirow}
\usepackage[normalem]{ulem}
\useunder{\uline}{\ul}{}

\usepackage{enumitem}

\usepackage{xparse}% http://ctan.org/pkg/xparse

\DeclareGraphicsExtensions{.pdf,.png}

\SetKwInput{KwInput}{Input}

\usepackage{longtable}
\usepackage{pgfplots}

\NewDocumentCommand{\var}{O{s} m O{}}{%
  \ensuremath{#1_{#2}^{#3}}% add \vphantom{<bizarre sup>}
}
\usepackage{siunitx}

\newcommand{\commentout}[1]{}

\definecolor{light-gray}{gray}{0.80}

\newcommand\appref{Appendix~\ref}
\newcommand\eref{Eq.~\ref}
\newcommand\fref{Figure~\ref}
\newcommand\tref{Table~\ref}

\newcommand\ha{ \rowcolor{orange!0}}

\newcommand\hc{ \rowcolor{orange!40}}

\def\0{{\bf 0}}

\usepackage{amsthm}

\usepackage{amsthm}

\newtheorem{mytheorem}{Theorem}
\newtheorem{assumption}[mytheorem]{Assumption}
\newtheorem{lemma}[mytheorem]{Lemma}
\newtheorem{remk}[mytheorem]{Remark}

\newcommand{\OURS}{\textsc{ANODEV2}\xspace}
\newcommand{\PREV}{\textsc{ANODE}\xspace}

\usepackage{fancyhdr}
\title{ANODEV2: A Coupled Neural ODE Evolution Framework}

\author{
Tianjun Zhang$^{1}$\thanks{Equal contribution.}  \and Zhewei Yao$^{1*}$ \and Amir Gholami$^{1*}$ \\
 Kurt Keutzer$^1$ \and Joseph Gonzalez$^1$ \and George Biros$^{2}$ \and Michael W. Mahoney$^{1,3}$ \\
$^1$University of California at Berkeley, $^2$University of Texas at Austin, $^3$ICSI\\
\{tianjunz, zheweiy, amirgh,  keutzer, jegonzal, and mahoneymw\}@berkeley.edu, biros@ices.utexas.edu
}

\begin{document}

\maketitle
\thispagestyle{empty}

%%%%%%%% BODY TEXT

\begin{abstract}
It has been observed that residual networks can be viewed as the explicit Euler discretization of an Ordinary Differential Equation (ODE). 
This observation has motivated the so-called Neural ODE models,
where it is possible to use higher order discretization schemes with the possibility of adaptive time stepping.
Here, we propose \OURS, which is an extension of the approach that also allows evolution of the neural
network parameters, in a coupled ODE-based formulation. The Neural ODE method introduced earlier is in fact  a special case of this new more general framework, which allows the evolution of the neural network's parameters in time, along with the activations.

We present the formulation of \OURS, derive the optimality conditions, and implement a coupled reaction-diffusion-advection version of this framework in PyTorch. 
We present empirical results using several different configurations of \OURS, testing them on multiple
models on CIFAR-10. 
We report results showing that this coupled ODE-based framework is indeed trainable, and that it achieves higher accuracy, as compared to the baseline models as well as the recently-proposed Neural ODE~approach.
\end{abstract}

\section{Introduction}
\label{sec:intro}
Residual networks~\cite{he2016deep,he2016identity} have enabled training of very deep neural networks (DNNs).
Recent work has shown an interesting connection between residual blocks and Ordinary Differential Equations (ODEs), showing that a residual network can be viewed as a discretization of a continuous ODE operator~\cite{weinan2017proposal,haber2017stable,ruthotto2018deep,lu2018beyond,ciccone2018nais,chen2018neural}. 
These formulations are commonly called \emph{Neural ODEs}, and here we follow the same convention. 
Neural ODEs provide a general framework that connects discrete DNNs to continuous
dynamical systems theory, as well as discretization and optimal control of ODEs, all subjects with very rich theory.

A basic Neural ODE formulation and its connection to residual networks (for a single block in a network) is the~following:
\begin{subequations}
\begin{align}
        z_1  &= z_0 + f(z_0,\theta) &&\quad \mbox{\small ResNet}, \\
        z_1 &= z_0 + \int_0^1 f(z(t),\theta) dt && \quad \mbox{\small ODE},\label{e:ode} \\
        z_1 &= z_0 + f(z_0,\theta) && \quad \mbox{\small ODE forward Euler}.
\end{align}
\end{subequations}
Here, $z_0$ is the input to the network and $z_1$ is the output activation; 
$\theta$ is the vector of network weights (independent of time); and $f(z, \theta)$ is the nonlinear operator defined by this block.  (Here, we have written the ODE $dz/dt = f(z,\theta)$ in terms of its solution at $t=1$.) We can see that a single-step of forward Euler discretization of the ODE is identical to a traditional residual block.  
Alternatively, we could use a different time-stepping scheme or, more interestingly, use more time steps.  
Once the connection to ODEs was identified, several groups have incorporated the Neural ODE structure in neural networks and evaluated its performance on several different learning tasks.  

A major challenge with training Neural ODEs is that backpropogating through ODE layers requires storage of all the intermediate activations (i.e. $z$) in time.  
In principle, the memory footprint of ODE layers has a cost of $\mathcal{O}(N_t)$ (where $N_t$ is the number of time steps to solve the ODE layer), which is prohibitive.  
The recent work of~\cite{chen2018neural} proposed an adjoint based method to address this, with a training strategy that required only storage of the activation at the end of the ODE layer.  
All the intermediate activations were then ``re-computed'' by solving the ODE layers backwards.  
However, it has been recently shown that such an approach could lead to incorrect gradients, due both to numerical instability of backward ODE solve, and also to inconsistencies that relate to optimizing infinite dimensional operators~\cite{gholami2019anode}.
This is basically the well-known Discretize-Then-Optimize (DTO) versus Optimize-Then-Discretize issue. The checkpointing-based method \PREV~\cite{gholami2019anode} was proposed to solve the incorrect gradient problem of Neural ODE.
More importantly, it was observed that using other discretization
schemes such as RK2 or RK4, or using more time steps, does not affect the
generalization performance of Neural ODEs as compared to baseline networks, despite the common belief~\cite{gholami2019anode}. This is a very important challenge, as lack of any performance gain obviates the need for using Neural ODEs.
In this paper, building on the latter approach of~\cite{gholami2019anode}, we propose \OURS, a more general Neural ODE framework that addresses this problem.
The key idea of \OURS is that it allows the evolution of \emph{both weights and activations} with a coupled system of ODEs:

\begin{equation}
    \begin{cases}
    z(1) = z(0) + \int_0^1 f(z(t), \theta(t)) dt ~~~~~~~~~~~~~~ \hfill \mbox{``parent network'',}\\
    \theta(t) = \theta(0) + \int_0^t q(\theta(t),p) dt,~~~\theta(0)=\theta_0~~~ \hfill \mbox{``weight network''}.\label{eq:simple_coupled}
    \end{cases}
\end{equation}
Here, $q(\cdot)$ is a nonlinear operator (essentially controlling the dynamics of the network parameters in time); $\theta_0$ and $p$ are the corresponding parameters for the weight network. 
Our approach allows $\theta$ to be time dependent: $\theta(t)$ is parameterized by the learnable dynamics of $d\theta/dt = q(\theta(t),p)$.
This, in turn, is parameterized by $\theta_0$ and $p$. 
In other words, instead of optimizing for a constant $\theta$, we optimize for $\theta_0$ and $p$. 
During inference, \emph{both} weights $\theta(t)$ and activations $z(t)$ are forward-propagated in time by solving~\eref{eq:simple_coupled}. 
Observe that if we set $q=0$ then we recover the Neural ODE approach proposed by~\cite{weinan2017proposal,haber2017stable,ruthotto2018deep,lu2018beyond,ciccone2018nais,chen2018neural}. \eref{eq:simple_coupled} replaces the problem of designing appropriate neural network blocks ($f$) with the problem of choosing appropriate function ($q$) in an ODE to model the changes of parameter $\theta$ (the weight network).

In summary, our main contributions are the following.
\begin{itemize}
    \item 
    We provide a general framework that extends Neural ODEs to a system of coupled ODEs which allows the coupled evolution of both model parameters and activations. 
    This coupled formulation addresses the challenge with Neural ODEs, in that using more time steps or different discretization schemes do not affect model's generalization performance~\cite{gholami2019anode}.
    \item 
    We derive the optimality conditions for how backpropagation should be performed for the coupled ODE formulation, using the so-called Karush–Kuhn–Tucker (KKT) conditions.
    In particular, we implement the corresponding DTO approach, along with a checkpointing scheme presented in~\cite{gholami2019anode}.
    \item 
    We test the framework using multiple different residual models on CIFAR-10 by
    considering different coupled formulations. 
    In particular, we show examples illustrating how a bio-physically motivated reaction-diffusion-advection (RDA) ODE could be used to model the evolution of the neural network~parameters.
\end{itemize}

Our work fits into a rich literature on neural evolution research~\cite{lindenmayer1968mathematical,turing1990chemical,belew1993evolving,bentley1999three,dellaert1996developmental,eggenberger1997evolving,hornby2002creating}. 
For example, several approaches similar to \OURS have been taken in the line of evolutionary computing, where an auxiliary ``child'' network is used to generate the parameters for a ``parent'' network. 
This approach permits the restriction of the effective depth that the activations must go through, since the parent network could have smaller weight space than the child network. 
One example is  HyperNEAT~\cite{stanley2009hypercube}, which uses ``Compositional Pattern Producing Networks'' (CPRNs) to evolve the model parameters~\cite{stanley2006exploiting,stanley2007compositional}. 
A similar approach using ``Compressed Weight Search'' was proposed in~\cite{koutnik2010evolving}.
A follow up work extended this approach by using differentiable CPRNs~\cite{fernando2016convolution}.
The authors show that neural network parameters could be encoded through a fully connected architecture.
Another seminal work in this direction is~\cite{schmidhuber1992learning,schmidhuber1993self}, where an auxiliary network
learns to produce ``context-aware'' weights in a recurrent neural network model.

A similar recent approach is taken in Hypernetworks~\cite{ha2016hypernetworks}, where model parameters are evolved through
an auxiliary learnable neural network. This approach is a special case of \OURS, which could be derived by using a single time step discretization of~\eref{eq:simple_coupled}, with a neural network for the evolution operator
(denoted by $q$ and introduced in the next section).

Our framework is a generalization of these evolutionary algorithms, and it provides more flexibility for modeling the evolution of the model parameters in time. 
For instance, we will show how RDA operators could be used for the evolution operator $q$, with negligible increase in the model parameter size.

\section{Methodology}
\label{sec:methodology}
In this section, we  discuss the formulation for the coupled ODE-based neural network model described above, and we derive the corresponding optimality conditions.  
For a typical learning problem, the goal is to minimize the empirical risk over a set of training examples. Given a loss function $\ell(\cdot)$, we seek to find weights, $\theta$, such that:
\begin{equation}\label{eq:basic_sum_obj}
\min_{\theta} \frac{1}{N} \sum_{i=1}^{N} \ell( z(\theta; x_i, y_i) ) + R(\theta),
\end{equation}
where
$R$ is a regularization operator, $(x_i,y_i)$ is the $i^{th}$ training sample and its label, and $N$ the number of training samples.
The loss function depends implicitly on $\theta$ through the network activation vector $z$. This problem is typically solved using Stochastic Gradient Descent (SGD) and backpropagation to compute the gradient of $z$ with respect to $\theta$.

\subsection{Neural ODE}

Consider the following notation for a residual block: $z_1 = z_0 + f(z_0; \theta)$, where $z_0$ is the input activation, $f(\cdot)$ is the neural network kernel (e.g., comprising a series of convolutional blocks with non-linear or linear activation functions), and $z_1$ is the output activation. 
As discussed above, an alternative view of a residual network is the following continuous-time formulation:
   $\frac{dz}{dt} = f(z(t); \theta)$,
with $z(t=0)=z_0$ (we will use both $z(t)$ and $z_t$ to denote activation at time $t$).
In the ODE-based formulation, this neural network has a continuous depth. 
In this case, we need to solve the following constrained optimization problem (Neural ODE):
\begin{equation}\label{eq:anodev1_loss}
\min_{\theta} \frac{1}{N} \sum_{i=1}^{N} \ell(z_1; x_i, y_i) + R(\theta) \quad
    \text{subject to:} \quad \frac{dz}{dt} = f(z(t),~~\theta), \quad z(0)=z_0.
\end{equation}

Note that in this formulation the neural network parameters do not change with respect to time. 
In fact, it has been observed that using adaptive time stepping or higher order discretization methods
such as Runge-Kutta does \emph{not} result in any gains in generalization performance using
the above framework~\cite{gholami2019anode}.
To address this, we extend the Neural ODEs by considering a system of coupled ODEs, where both the model parameters and the activations evolve in time. 
This formulation is slightly more general than what we described in the introduction. 
We introduce an auxiliary dynamical system for $w(t)$, which we use to define $\theta$.
This allows for more general evolution of the model parameters and activations, which will be discussed in~\S\ref{sec:two_scheme}.
In particular, we propose the following formulation:
\begin{subequations}
\label{eq:anodev2_loss}
\begin{align}
\min_{p,w_0} \gJ(z(1)) &= \frac{1}{N} \sum_{i=1}^{N} \ell(z(1); x_i, y_i) + R(w_0, p),\\
    \frac{dz}{dt} &= f(z(t), \theta(t)),\ z(0)=z_0\quad~~~~~~~~ \hfill \mbox{``Activation ODE''}, \label{eq:activation_ode}\\
    \frac{\partial w}{\partial t} &= q(w; p),\ w(0)=w_0\quad  ~~~~~~~~~~~~~\hfill \mbox{``Evolution ODE''},\label{eq:anodev2_loss_w_pde}\\
    \theta(t) &= \int_0^t K(t-\tau)w(\tau)d\tau\label{eq:anodev2_theta_formula}.
\end{align}
\end{subequations}
Note that here $\theta(t)$ is a function of time, and it is parameterized by the whole dynamics of $w(t)$ and a time convolution kernel $K$ (which in the simplest form could be a Dirac delta function so that $\theta(t) = w(t)$). 
Also,  $q(w,p)$ can be a general function, e.g., another neural network, a linear operator, or even a discretized Partial Differential Equation (PDE) based operator. 
The latter is useful if we consider the $\theta(t)$ as a function $\theta(u,t)$, where $u$ parameterizes the signal space (e.g., 2D pixel space for images).  
This formulation allows for rich variations of $\theta(t)$,  while using a lower dimensional parameterization: notice that implicitly we have that $\theta(t) = \theta(w_0, p,t)$. 
Also, this formulation permits novel regularization techniques. 
For instance, instead of regularizing $\theta(t)$, we can regularize $w_0$ and $p$. 

A crucial question here  is: how should one perform backpropagation for this formulation? 
It is instructive to compute the actual derivatives to illustrate the structure of the problem.
To derive the optimality conditions for this constrained optimization problem, we need first to form
the Lagrangian operator, and then we derive the KKT conditions.
Here is the Lagrangian:

\begin{align}
\begin{split}
    \gL &= \gJ(z_1) + \int_0^1\alpha(t) \cdot \left(\frac{dz}{dt} - f(z(t),\theta(t))\right)dt 
    +\int_0^1\beta(t)\cdot \left(\frac{\partial w}{\partial t} - q(w; p)\right)dt \\
    &+\int_0^1\gamma(t) \cdot \left(\theta(t) - \int_0^t K(t-\tau)w(\tau)d\tau\right)dt.
\end{split}
\end{align}
Here, $\alpha(t)$, $\beta(t)$, and $\gamma(t)$ are the corresponding adjoint variables (Lagrange multiplier vector functions) for the constraints in~\eref{eq:anodev2_loss}. The solution to the optimization
problem of~\eref{eq:anodev2_loss} could be found by computing the stationary points of the Lagrangian (the KKT conditions), which are the gradient of $\gL$ with respect to $z(t), w(t), \theta(t), p, w_0$ and the adjoints $ \alpha(t),\beta(t),\gamma(t)$.  The variations of $\gL$ with respect to the three adjoint functions just result in the ODE constraints in~\eref{eq:anodev2_loss}. The remaining variations of $\gL$ are the most interesting and are given below
(please see \appref{sec:opt_condtion} for the derivation):

\begin{subequations}
\label{eq:ktt_conditinos}
\begin{align}
\frac{\partial \gJ(z_1)}{\partial z_1} + \alpha_1=0, ~~~~ -\frac{\partial \alpha}{\partial t} - \left(\frac{\partial f}{\partial z}\right)^T \alpha(t) = 0;\qquad (\partial \gL_z)\label{eq:kkt_conditions_adjoint1}\\
-\left(\frac{\partial f}{\partial \theta}\right)^T\alpha(t)  +\gamma(t) = 0;
\qquad \hfill (\partial \gL_\theta)\label{eq:kkt_conditions_inversion2}\\
-\frac{\partial \beta(t)}{\partial t} - \left(\frac{\partial q}{\partial w}\right)^T\!\!\!\beta(t) - \int_t^1 K^T(\tau-t)\gamma(\tau) d\tau = 0 ,~~\beta(1) = 0; \qquad \hfill (\partial \gL_w)\label{eq:kkt_conditions_inversion3}\\
-\beta(0) + \frac{\partial R}{\partial w_0}= g_{w_0}; \qquad \hfill (\partial \gL_{w_0}) \label{eq:kkt_conditions_inversion4}\\
\frac{\partial R}{\partial p} - \int_0^1 \left(\frac{\partial q}{\partial p}\right)^T \!\!\!\beta(t) dt = g_p.\qquad \hfill (\partial \gL_p) \label{eq:kkt_conditions_inversion1}
\end{align}
\end{subequations}
To compute the gradients $g_p$ and $g_{w_0}$ (\eref{eq:kkt_conditions_inversion4} and~\eref{eq:kkt_conditions_inversion1}), we proceed as follows.  Given $w_0$ and $p$, we forward propagate $w_0$ to compute $w(t)$
and then $\theta(t)$. Then, using $\theta(t)$ we can compute the activations $z(t)$.

Afterward, we need to
solve the first adjoint equation for $\alpha(t)$ using the terminal condition of $\alpha_1 = -\frac{\partial \gJ(z_1)}{\partial z_1}$.
Having $\alpha(t)$ we can compute the second adjoint variable $\gamma(t)$ from~\eref{eq:kkt_conditions_inversion2}.
Lastly, we need to plug in $\gamma(t)$ into~\eref{eq:kkt_conditions_inversion3} to solve for $\beta(t)$ which is the term needed to compute the
gradients (by plugging it in in~\eref{eq:kkt_conditions_inversion4},~\eref{eq:kkt_conditions_inversion1}).
We use the Discretize-Then-Optimize (DTO) method to find the gradients~\cite{gholami2019anode}.

Notice that if we set $q=0$ then we will derive the optimality conditions for the Neural ODE without any dynamics for the model parameters, which was the model presented in~\cite{chen2018neural}. The benefit of our more general framework is that we can encapsulate the time dynamics of the model parameter without increasing
the memory footprint of the model. In fact, this approach only requires
storing initial conditions for the parameters, which are 
parameterized by $w_0$, along with the parameters of the control
operator $q$, which are denoted by $p$. As we show in the results
section (\S\ref{sec:results}) the latter have negligible memory footprint, but yet
allow rich representation of model parameter dynamics.

{\bf PDE-inspired formulation.} 
There are several different models one could consider for the $q(w, p)$, the evolution function for the neural network parameters. One possibility is to use an auxiliary neural network such as the approach used in HyperNetworks~\cite{ha2016hypernetworks}.
However, this may increase the total number of parameters. Inspired by Turing's reaction-diffusion-advection partial differential equation models for pattern formation, we view a convolutional filter as a time-varying pattern, where the NN parameters evolve in time~\cite{turing1990chemical}. To illustrate this, we consider a PDE-based model for the control
operator $q$, as follows:
\begin{equation}\label{eq:w_rda_pde}
    \frac{dw}{dt} = \sigma(d \Delta w + \upsilon \cdot \nabla w + \rho  w),
\end{equation}
where $d$ is control diffusion ($\Delta w$), $\upsilon$ controls the advection/transport ($\nabla w$), $\rho$ controls the reaction ($w$), and $\sigma$ is a nonlinear activation (such as sigmoid or tanh). 
Here, we are viewing the weights $w$ as a time series signal, starting from the initial
signal, $w_0$. This initial condition is then evolved in time to produce $w_1$. In fact, one can show that
this formulation can evolve the initial parameters, $w_0$, to any arbitrary weights $w_1$, if there exists a 
diffeomorphic transformation of between the two distributions (i.e., if there  exists a velocity field $\upsilon$ such $w_1$ is the solution of~\eref{eq:w_rda_pde}, with initial condition $w_0$~\cite{younes2010shapes}).  

\begin{figure}[!htbp]
\centering
\includegraphics[width=.8\textwidth]{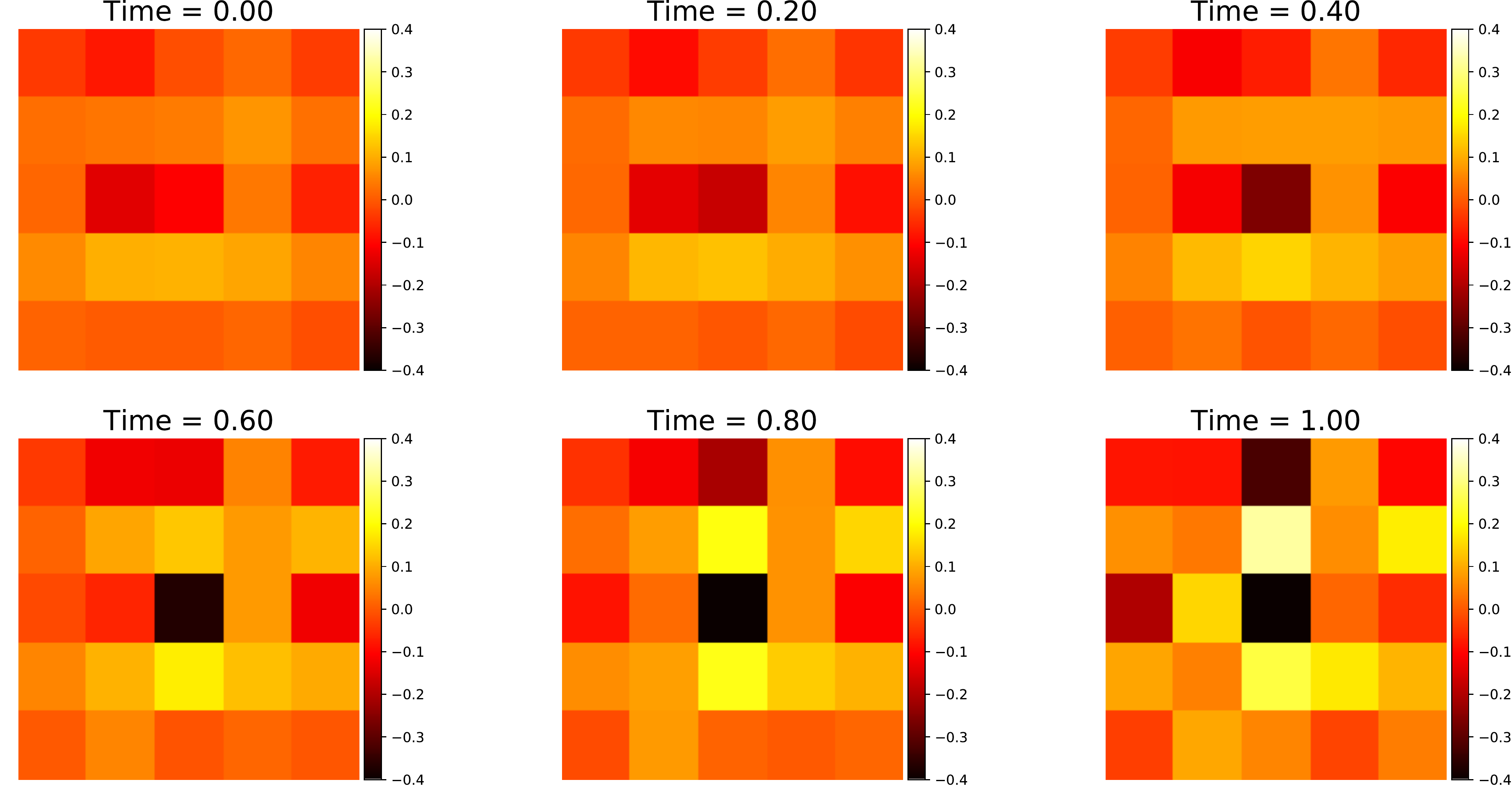}
\caption{
Illustration of how different convolutional operators are evolved in
time during the coupled neural ODE solve (through the evolution operator $q$ in~\eref{eq:anodev2_loss_w_pde}).
The figure corresponds to the first channel of the first convolution kernel parameters of AlexNet.
These filters will be applied to activation in different time steps (through the $f$ operator in the coupled formulation in~\eref{eq:activation_ode}).
This is schematically shown in~\fref{fig:anode} for three of the filters (the filters are denoted by different shades of brown bars denoted by $\theta$).
% We can see that at the beginning, the variance of the parameter are quite small.
As the time step increases, the kernel turns out to focus on some specific part on the activation map. 
Similar illustrations for ResNet-4 and ResNet-10 are shown in~\fref{fig:resnet4_kernel} and~\ref{fig:resnet10_kernel} in the appendix.
}
  \label{fig:qalexnet_kernel}
\end{figure}

Although this operator is mainly used as an example control block (i.e., \OURS is not limited to this model), the
RDA operator can capture interesting dynamics for model parameter evolution. 
For instance, consider a  Gaussian operator for a convolutional kernel with unit variance as shown in~\fref{fig:RDA_example}.
A diffusion operator can simulate multiple different Gaussian distributions with different variances in time. This requires storing only a single diffusion parameter (i.e., the diffusion parmaeter $d$ in~\eref{eq:w_rda_pde}). 
Another interesting operator is the advection/transport operator which models species transport. 
For the Gaussian case, this operator could transport the center of the Gaussian to different positions other than the center, as shown in second row of~\fref{fig:RDA_example}. 
Finally, the reaction operator could allow growth or decay of the intensity
of the convolution filters (third row of~\fref{fig:RDA_example}). 
The full RDA operator could encapsulate more complex dynamics of the neural network parameters in time. 
An synthetic example is shown in~\fref{fig:RDA_example} in the appendix, and
a real example ($5\times5$ convolutional kernel of AlexNet) is shown in~\fref{fig:qalexnet_kernel}.

% ------------------------------------------------------------
% ------------------------------------------------------------

\subsection{Two methods used in this paper}\label{sec:two_scheme}
% \vspace{-2mm}
The motivation to introduce the auxiliary variable $w$ in~\eref{eq:anodev2_loss} is to enable different coupling configurations
between model parameters, $\theta$, and activations, $z$. We use two different coupling configurations of \OURS as described below.

\paragraph{Configuration 1} 
% \vspace{-2mm}
Here, we use multiple time steps to solve for  both $z$ and $\theta$ in the network, instead of just
one time step as in the original ResNet.
In this setting, we will alternatively update the value of $z$ and $\theta$ according to the following equation (for details, see~\appref{sec:rda_example}): 
% Then, the discretized solution of~\eref{eq:w_rda_pde} will be as follows (for details, see~\appref{sec:rda_example}):  
\begin{equation}\label{eq:nonlinear_solution}
% z_{t_0+\delta t} = z_{t_0} + \delta tf(z_{t_0}; \theta_{t_0});~~~\theta_{t_0+\delta t} =\sigma\Big( F^{-1}\big( F(\delta t\tau  \Delta + \delta t\upsilon \cdot \nabla + \delta t\rho) \odot F(\theta_{t_0}) \big) \Big),
z_{t_0+\delta t} = z_{t_0} + \delta tf(z_{t_0}; \theta_{t_0});~~~\theta_{t_0+\delta t} =\sigma\left(   F^{-1}\left( \exp((-d k^2 + i k\upsilon + \rho)\delta t) F(\theta_{t_0})\right)    \right).
\end{equation}
where $\delta t$ is the discretization time scale, and F is Fast Fourier Transform (FFT) operator.
where $t_0 \in \{0, 1/5, 2/5, 3/5, 4/5\}$ in the forward solver. 

If we use $N_t$ time steps to solve the equation, the computational cost for an ODE block will be roughly $N_t$ times more expensive, compared to that for the original residual block (note the approach presented in~\cite{chen2018neural} also increases computational cost by the same $N_t$ factor).
This network can be viewed as applying $N_t$ different residual blocks in the network but with
neural network weights that evolve in time.
Note that this configuration does not increase the parameter size of the original network, except for a slight overhead of $d,~\upsilon$ and $\rho$.

\begin{figure}[!htbp]
\centering
\includegraphics[width=0.87\textwidth]{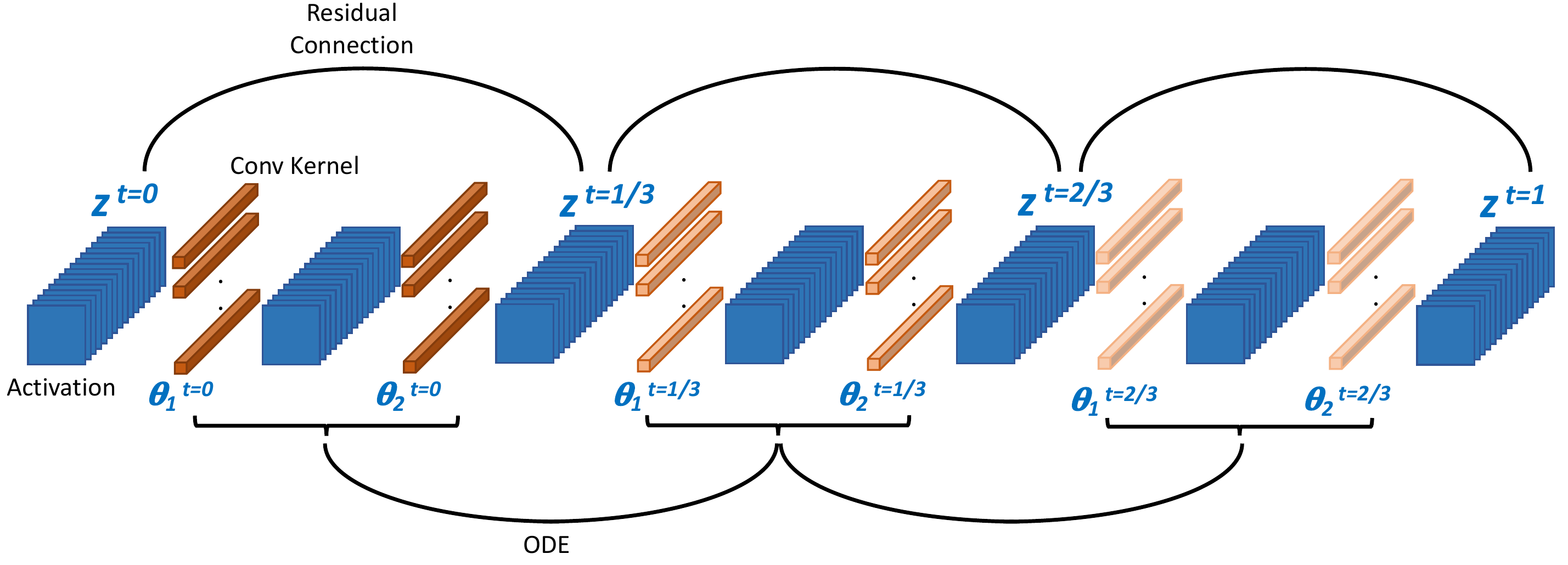}\\
\includegraphics[trim=0 0 0 1.1cm, clip,width=0.87\textwidth]{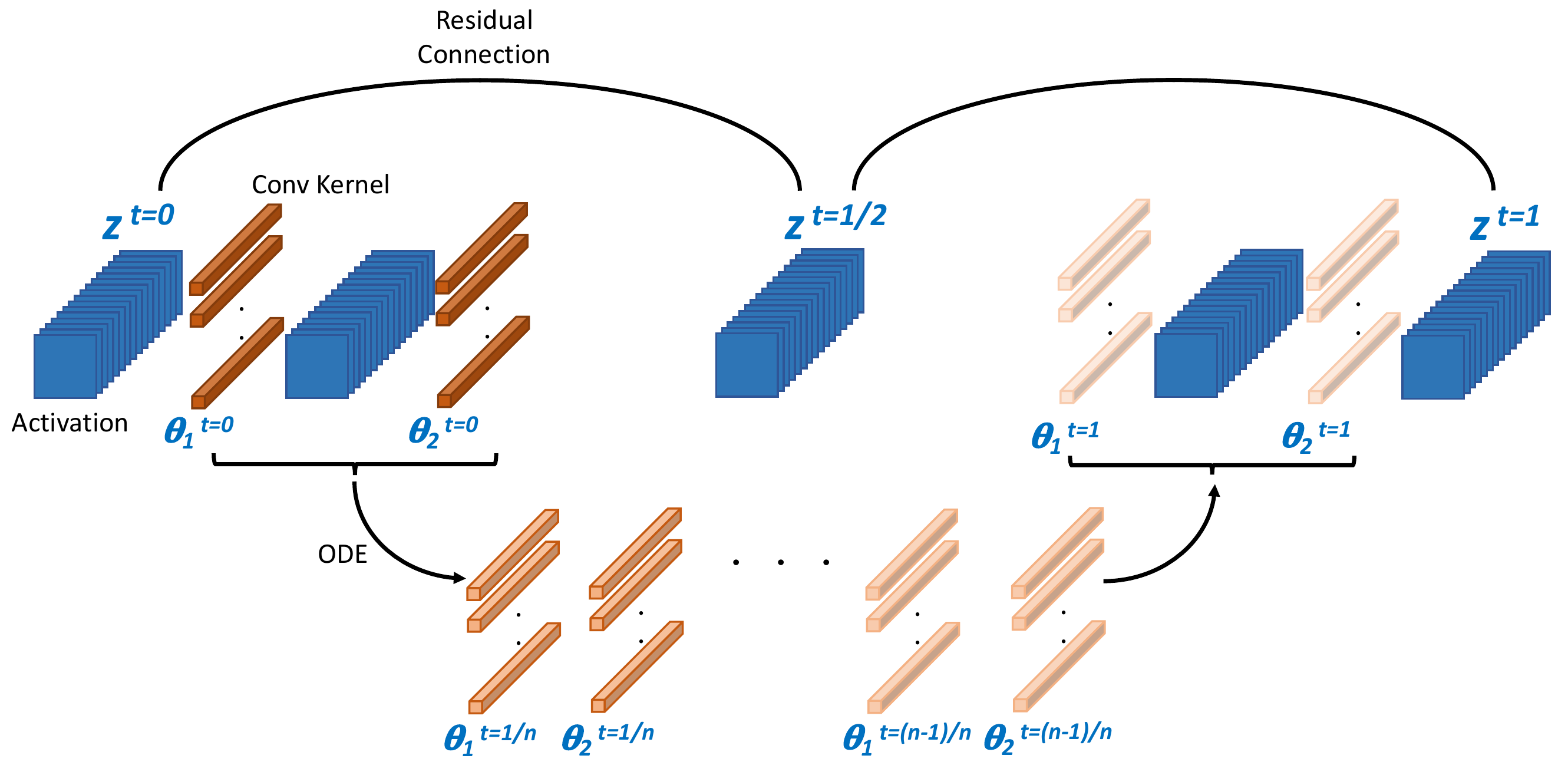}
\caption{
Illustration of different configurations in \OURS. The top figure shows configuration 1, where both the activation $z$ and weights $\theta$ are evolved through a coupled system of ODEs. 
During inference, we solve both of these ODEs forward in time.
Blue boxes in the figure represent activation with multiple channels; all the bars (in different shades of brown) represent the convolution kernel evolving in time.
The convolution weights $\theta$ are computed by solving an auxiliary ODE. That is, at every time step, we solve both z and $\theta$ forward concurrently.
The bottom figure shows configuration 2, where first the weights are evolved in time before
applying them to the activations. Comparing to configuration 1, only the first and last weights are applied. 
}
  \label{fig:anode}
\end{figure}

The configuration 1 is shown in~\fref{fig:anode} (top), where the model parameters
and activations are solved with the same discretization scheme. This is similar to the Neural ODE framework of~\cite{chen2018neural},
except that the model parameters are evolved in time, whereas in~\cite{chen2018neural} the same model parameters are applied to the activations (and only time horizon is changed). The dynamics of the model parameters are illustrated by different colors used for the convolution kernels in top of~\fref{fig:anode}. This configuration is equivalent to using the Dirac delta function for the $K$ function in~\eref{eq:anodev2_theta_formula}. 

In short, this configuration allows evolution of both the model parameters and activations,
but both have the same time discretization. While this is a simple configuration, but it is not very general, since the model parameters may require more time to evolve and prematurely applying them
to input activation may not be optimal. The next configuration relaxes this constraint to addresses this limitation.

\paragraph{Configuration 2} 
% \vspace{-2mm}
%\zhewei{This paragraph is messy. First half and second half are talking about the same thing}
\OURS supports different coupling configurations between the dynamics of activations
and model parameters. 
For example, it is possible to not restrict the dynamics of $\theta$ and  $z$ to align in time,
which is the second configuration that we consider. Here, we allow model parameters to evolve and only apply to activations after a fixed number of time steps. 
For instance, consider the Gaussian example illustrated in~\fref{fig:RDA_example} in~\appref{sec:rda_sim}. In configuration 1, a residual block is created for each of the three time steps. However, in configuration 2, we only apply the first and last time evolutions of the parameters (i.e., we only use $w_0$ and $w_1$ to apply to activations as shown in~\fref{fig:anode}). This configuration allows sufficient time for the model parameters to evolve, and importantly it limits the depth of the network that activations go through.
In this case, the depth of the network is increased by a factor of two, instead of $N_t$
as in configuration 1 (which is the approach used in~\cite{chen2018neural,gholami2019anode}).

Both configuration 1 and configuration 2 are supported in \OURS, and we will present preliminary results for both settings.

% We experimented with these 2 configurations on AlexNet, ResNet-4 and ResNet-10. The results will be discussed in Section \ref{sec:results}.

% ------------------------------------------------------------
% ------------------------------------------------------------

\section{Results}
\label{sec:results}
In this section, we report the results of \OURS
for the two configurations discussed in~\secref{sec:methodology},
on the CIFAR-10 dataset, which consists of 60,000 32$\times$32 colour images in 10 classes.
The framework is developed as a library in Pytorch, and it uses the checkpointing method proposed in~\cite{gholami2019anode}, along with the DTO formulation of the optimality conditions shown in~\eref{eq:ktt_conditinos}.

We test \OURS on AlexNet with residual connections, as well as two different ResNets.
See~\appref{sec:rda_example} and~\appref{sec:model_arch} for the details of training settings and model architectures.
We consider the two coupling configurations between
the evolution of the activations and model parameters, as discussed next.

\subsection{Configuration 1}
We first start with configuration 1, which is the same as the setting
used in~\cite{chen2018neural,gholami2019anode}.
The model parameters and activations are evolved in time concurrently and for each time step, as shown in~\fref{fig:anode} (top).
All the experiments were repeated five times, and we report both the min/max accuracy as well as the average of these five runs.
The results are shown in~\tref{tab:Nt_5_results}.

Note that the coupled ODE based approach outperforms the baseline in all three of the statistical properties (i.e.,~min/max/average accuracy). 
For example, on ResNet-10 the coupled ODE network achieves 89.04\% average test accuracy, as compared to 88.10\% of baseline, which is 0.94\% better.
Meanwhile, a noticeable observation is that the minimum performance of the coupled ODE based network is comparable or even better than the maximum performance of baseline. 
The coupled ODE based AlexNet has 88.59\% minimum accuracy which is 1.44\% higher than the best performance of baseline out of five runs.
Hence, the generalization performances of the coupled ODE based network are consistently better than those of the baseline.
It is important to note that the model parameter size of the coupled ODE approach in \OURS is the same as that of the baseline.
This is because the size of the evolution parameters $p$ in~\eref{eq:anodev2_loss} is negligible (please see~\tref{tab:parameter_size}).

\begin{table*}[!htbp]
\caption{Results for using $N_t=5$ time steps to solve $z$ and $\theta$ in a neural network with configuration 1. We tested on AlexNet, ResNet-4, and ResNet-10. We get 1.75\%, 1.16\% and 0.94\% improvement over the baseline respectively. Note that the model size of \OURS and the baseline are comparable. }
\small
\setlength\tabcolsep{3.5pt}
\label{tab:Nt_5_results}
\centering
\begin{tabular}{lcccccccccccc} \toprule
                      & \multicolumn{2}{c}{AlexNet}  &  \multicolumn{2}{c}{ResNet-4} &   \multicolumn{2}{c}{ResNet-10} \\ 
                        \cmidrule{2-3}                  \cmidrule{4-5}              \cmidrule{6-7}                  
    & {Min / Max} & {Avg}    & {Min / Max} & {Avg}  & {Min / Max} & {Avg}   \\
    \midrule
\hc Baseline     &86.84\% / 87.15\% &87.03\% &76.47\% / 77.35\% &76.95\%   &87.79\% / 88.52\% &88.10\% \\
\ha \OURS        &\textbf{88.59\%} / \textbf{88.96\%} & \textbf{88.78\%} &\textbf{77.27\%} / \textbf{78.58\%} &\textbf{78.11\%}   &\textbf{88.67\%} / \textbf{89.39\%} &\textbf{89.04\%} \\
\hc Imp.         &1.75\%  / 1.81\%  &1.75\%  &0.80\%  / 1.23\%  &1.16\%    &0.88\%  / 0.87\%  &0.94\%  \\
     \bottomrule 
\end{tabular}
\end{table*}

The dynamics of how the neural network parameters evolve in this configuration are illustrated in~\fref{fig:qalexnet_kernel},
where we extract the first $5\times5$ convolution of AlexNet and show how it evolves in time.
Here, $Time$ represents how long $\theta$ evolves in time, i.e., $Time=0$ shows the result of $\theta(t=0)$ and $Time=1$ shows the result of $\theta(t=1)$.
It can be clearly seen that the coupled ODE based method encapsulates more complex dynamics of $\theta$ in time.
Similar illustrations for ResNet-4 and ResNet-10 are shown in~\fref{fig:resnet4_kernel} and~\ref{fig:resnet10_kernel} in~\appref{sec:model_arch}.

\subsection{Configuration 2}
Here, we test configuration 2, where the evolution of the parameters and the activations
could have different time steps. This means the parameter is applied only after a certain number of time
steps of evolution, i.e., not at every time step, which was the case in configuration 1. 
This effectively reduces
the network depth and the computational cost, and it
allows sufficient time for the neurons to be evolved, instead of naively applying them at each
time step.
An illustration for this configuration is shown in~\fref{fig:anode} (bottom). 
The results on AlexNet, ResNet4, and ResNet10 are shown in~\tref{tab:Nt_10_theta_results}, where
we again report the min/max/average accuracy over five runs.
As in the previous setting (configuration 1), here in configuration 2, the coupled ODE based network performs better in all cases. 
The minimum performance of the coupled ODE based network still is comparable or even better than the maximum performance of the baseline. 
Although the overall performance of this configuration 2 is slightly worse than that of configuration 1, the computational cost is much less, due to the
smaller effective depth of the network that the activations go through.

\begin{table*}[!htbp]
\caption{Results for using $N_t=2$ time steps to solve $z$ in neural network and $N_t=10$ to solve $\theta$ in the ODE block (configuration 2). \OURS achieves $1.23\%$, $0.78\%$ and 0.83\% improvement over the baseline respectively. Note that the model size is comparable to the baseline in~\tref{tab:Nt_5_results}.}
\footnotesize
\setlength\tabcolsep{3.5pt}
\label{tab:Nt_10_theta_results}
\centering
\begin{tabular}{lcccccccccccc} \toprule
                      & \multicolumn{2}{c}{AlexNet}  &  \multicolumn{2}{c}{ResNet-4} &   \multicolumn{2}{c}{ResNet-10} \\ 
                        \cmidrule{2-3}                  \cmidrule{4-5}              \cmidrule{6-7}                  
    & {Min / Max} & {Avg}    & {Min / Max} & {Avg}  & {Min / Max} & {Avg}   \\
    \midrule
\hc Baseline     &86.84\% / 87.15\% &87.03\% &76.47\%  / 77.35\% &76.95\%    & 87.79\% / 88.52\% &88.10\% \\
\ha \OURS        &\textbf{88.1\%}  / \textbf{88.33\%} &\textbf{88.26\%} & \textbf{77.23\%} / \textbf{78.28\%}  & \textbf{77.73\%}  & \textbf{88.65\%} / \textbf{89.19\%} &\textbf{88.93\%} \\
\hc Imp.         &1.26\%  / 1.18\%  &1.23\%  &0.76\%   / 0.93\% & 0.78\%     & 0.86\%  / 0.67\%  & 0.83\% \\
     \bottomrule 
\end{tabular}
\end{table*}

\section{Ablation Study}\label{sec:ablation_study}
% \vspace{-2mm}
In this section, we compare \OURS to models with the same number of parameters. In particular,
we compare with the Neural ODE approach of~\cite{chen2018neural}. As mentioned before, the approach used
in this paper to save memory results in numerical instability, and to allow for fair comparison,
we also compare with ANODE presented in~\cite{gholami2019anode}, which addresses the instability problem.
Precisely, we use two time steps for the
activation ODE (\eref{eq:activation_ode}) and ten time steps for the evolution of the model parameters (\eref{eq:anodev2_loss_w_pde}). 
% In this setting, both the FLOPS and model sizes are roughly the same, allowing us
% to test the efficacy of evolving model parameters.
The results are shown in~\tref{tab:ablation_study}.

As one can see, there is indeed benefit in allowing the model parameters to evolve in time.
This is not surprising, since it gives more flexibility to the neural network to evolve the model parameters.
Also note that the performance of the Neural ODE approach used in~\cite{chen2018neural} is significantly worse than \PREV and \OURS.
The results remained the same despite hyper-parameter tuning. However, this is expected as the Neural ODE method in~\cite{chen2018neural}
may result in incorrect gradient information~\cite{gholami2019anode}. However, even addressing this instability with \PREV results
in suboptimal performance as compared to \OURS.
% The \PREV results are derived using the DTO approach with checkpointing presented in~\cite{gholami2019anode}.
Also note that
evolving model parameters has a negligible computational cost, since we can actually use analytical solutions
for solving the RDA, which is discussed in~\appref{sec:rda_example}.

We also present the parameter sizes of the both configurations of \OURS as well as the Neural ODE and ANODE models tested above.
Table~\ref{tab:parameter_size} summarizes all the results.
It can be clearly seen that the model sizes of both configurations are roughly the same as those of the baseline models. 
In fact, configuration 1 grows the parameter sizes of AlexNet, ResNet-4, and ResNet-10 by only $0.5\%$ to $6.7\%$, as compared to those of baseline models.
In configuration 2, the parameter size increases from $0.2\%$ to $3.6\%$ compared to baseline model. 
(Note that we even count the additional batch norm parameters for fair comparison.)

% Comparing with the ablation network used in~\secref{sec:ablation_study}, in which we apply the same model parameters for multiple time steps,
% \OURS configuration 2 has roughly the same number of parameters. 

% could lead to incorrect gradients. So for a fair comparison, we also test \PREV, which is a unconditionally accurate memory-efficient gradients Neural ODE.
% The parameter sizes are shown in~\tref{tab:parameter_size}, and results are shown in~\tref{tab:ablation_study}.
% \subsection{Accuracy Comparison}\label{sec:FLOPS_comp}
% For Neural ODE (which is the configuration used in~\cite{chen2018neural}) and \PREV~\cite{gholami2019anode}, we remove the evolution of the model parameters and fix them to stale values in time.
% First we start by comparing the generalization performance We compare the results with the case where the model parameters are evolved in time, which corresponds to the results of~\tref{tab:Nt_10_theta_results}.
% Precisely, we use two time steps for the
% activation ODE (\eref{eq:activation_ode}) and ten time steps for the evolution of the model parameters (\eref{eq:anodev2_loss_w_pde}). 
% In this setting, both the FLOPS and model sizes are roughly the same, allowing us
% to test the efficacy of evolving model parameters.
% The results are shown in~\tref{tab:ablation_study}.
\begin{table*}[!htbp]
\caption{Parameter comparison for two \OURS configurations, Neural ODE, \PREV, and the baseline network. The parameter size of \OURS is comparable with the others.}
\small
\label{tab:parameter_size}
\centering
\begin{tabular}{lcccccccccccc} \toprule
                    & AlexNet  & ResNet-4  &  ResNet-10  \\ 
                      
    \midrule
\hc Baseline        & 1756.68K & 7.71K & 44.19K     \\
\ha Neural ODE~\cite{chen2018neural}        & 1757.13K & 7.96K & 44.95K  \\
\hc \PREV~\cite{gholami2019anode}        & 1757.13K & 7.96K & 44.95K  \\
\ha \OURS conf. 1 & 1757.51K & 8.23K & 45.77K  \\
\hc \OURS conf. 2 & 1757.13K & 7.99K & 45.05K  \\
     \bottomrule 
\end{tabular}
\end{table*}

\begin{table*}[!htbp]
\caption{
We use $N_t=2$ time steps to solve for $z$ in the neural network and keep $\theta$ as static for Neural ODE and \PREV. We tested all configurations on AlexNet, ResNet-4 and ResNet-10. The results show that Neural ODE get significantly worse results, compared to \OURS (with Configuration 2) and \PREV. \OURS gets $0.24\%$, $0.43\%$ and $0.33\%$ improvement over \PREV, respectively. The model size comparison is shown in~\tref{tab:parameter_size}.}
\footnotesize
\setlength\tabcolsep{2.5pt}
\label{tab:ablation_study}
\centering
\begin{tabular}{lcccccccccccc} \toprule
                      & \multicolumn{2}{c}{AlexNet}  &  \multicolumn{2}{c}{ResNet-4} &   \multicolumn{2}{c}{ResNet-10} \\ 
                        \cmidrule{2-3}                  \cmidrule{4-5}              \cmidrule{6-7}                  
    & {Min / Max} & {Avg}    & {Min / Max} & {Avg}  & {Min / Max} & {Avg}   \\
    \midrule
\hc Baseline     &86.84\% / 87.15\% &87.03\% &76.47\% / 77.35\% &76.95\%   &87.79\%  / 88.52\% &88.10\% \\
\ha NeuralODE~\cite{chen2018neural}     &74.54\% / 76.78\% &75.67\% &44.73\% / 49.91\% &47.37\%   &64.7\%  / 70.06\% &67.94\% \\
\hc \PREV~\cite{gholami2019anode}  &87.86\% / 88.14\% &88.02\% &76.92\% / 77.45\% &77.30\% & 88.48\% / 88.75\% & 88.60\%  \\
\ha \OURS Conf. 2  &\textbf{88.1\%}  / \textbf{88.33\%} &\textbf{88.26\%} & \textbf{77.23\%} / \textbf{78.28\%}  & \textbf{77.73\%}  & \textbf{88.65\%} / \textbf{89.19\%} &\textbf{88.93\%} \\
     \bottomrule 
\end{tabular}
\end{table*}

% \subsection{Parameter Size Comparison}\label{sec:param_size_comp}

\section{Conclusions}
\label{sec:conclusions}
The connection between residual networks and ODEs has been discussed in recent work.
Here, motivated by work in neural evolution, we propose \OURS, which is a more general extension of this approach, obtained by introducing a coupled ODE based framework.
The framework allows dynamical evolution of both the residual parameters as well as the activations in a coupled 
ODE formulation. This provides more flexibility to the neural network to adjust the parameters to achieve better generalization performance.
We derived the optimality conditions for this coupled formulation and presented preliminary empirical results using
two different configurations, and we showed that we can indeed train such models using our differential framework.
The results on three Neural Networks (AlexNet, ResNet-4, and ResNet-10) all show
accuracy gains across five different trials. In fact, the worst accuracy of the coupled ODE formulation was better than the
best performance of the baseline. This is achieved with negligible change in the model parameter size.
To the best of the our knowledge, this is the first coupled ODE formulation that allows for the evolution of the
model parameters in time along with the activations. We are working on extending the framework for other learning tasks.
The source code will be released as open source software to the public.

\section*{Acknowledgments}
This work was supported by a gracious fund from Intel corporation,
Berkeley Deep Drive (BDD), and Berkeley AI Research (BAIR) sponsors.
We would like
to thank the Intel VLAB team for providing us with access to their computing cluster.
We also gratefully acknowledge the support of NVIDIA Corporation for their donation of two Titan Xp GPU used for this research.
MWM would also 
like to acknowledge ARO, DARPA, NSF, ONR, and Intel for providing partial support of this work.

\clearpage
\bibliographystyle{ieeetr}
{
\Large
\bibliography{ref.bib}}

\begin{thebibliography}{10}

\bibitem{he2016deep}
K.~He, X.~Zhang, S.~Ren, and J.~Sun, ``Deep residual learning for image
  recognition,'' in {\em Proceedings of the IEEE conference on computer vision
  and pattern recognition}, pp.~770--778, 2016.

\bibitem{he2016identity}
K.~He, X.~Zhang, S.~Ren, and J.~Sun, ``Identity mappings in deep residual
  networks,'' in {\em European conference on computer vision}, pp.~630--645,
  Springer, 2016.

\bibitem{weinan2017proposal}
E.~Weinan, ``A proposal on machine learning via dynamical systems,'' {\em
  Communications in Mathematics and Statistics}, vol.~5, no.~1, pp.~1--11,
  2017.

\bibitem{haber2017stable}
E.~Haber and L.~Ruthotto, ``Stable architectures for deep neural networks,''
  {\em Inverse Problems}, vol.~34, no.~1, p.~014004, 2017.

\bibitem{ruthotto2018deep}
L.~Ruthotto and E.~Haber, ``Deep neural networks motivated by partial
  differential equations,'' {\em arXiv preprint arXiv:1804.04272}, 2018.

\bibitem{lu2018beyond}
Y.~Lu, A.~Zhong, Q.~Li, and B.~Dong, ``Beyond finite layer neural networks:
  Bridging deep architectures and numerical differential equations,'' in {\em
  International Conference on Machine Learning}, pp.~3282--3291, 2018.

\bibitem{ciccone2018nais}
M.~Ciccone, M.~Gallieri, J.~Masci, C.~Osendorfer, and F.~Gomez, ``{NAIS-Net}:
  stable deep networks from non-autonomous differential equations,'' in {\em
  Advances in Neural Information Processing Systems}, pp.~3025--3035, 2018.

\bibitem{chen2018neural}
T.~Q. Chen, Y.~Rubanova, J.~Bettencourt, and D.~K. Duvenaud, ``Neural ordinary
  differential equations,'' in {\em Advances in Neural Information Processing
  Systems}, pp.~6571--6583, 2018.

\bibitem{gholami2019anode}
A.~Gholami, K.~Keutzer, and G.~Biros, ``{ANODE}: Unconditionally accurate
  memory-efficient gradients for neural odes,'' {\em 2019 International Joint
  Conference on Artificial Intelligence, (arXiv:1902.10298)}, 2019.

\bibitem{lindenmayer1968mathematical}
A.~Lindenmayer, ``Mathematical models for cellular interactions in development
  i. filaments with one-sided inputs,'' {\em Journal of theoretical biology},
  vol.~18, no.~3, pp.~280--299, 1968.

\bibitem{turing1990chemical}
A.~M. Turing, ``The chemical basis of morphogenesis,'' {\em Bulletin of
  mathematical biology}, vol.~52, no.~1-2, pp.~153--197, 1990.

\bibitem{belew1993evolving}
R.~K. Belew and T.~E. Kammeyer, ``Evolving aesthetic sorting networks using
  developmental grammars.,'' in {\em ICGA}, p.~629, Citeseer, 1993.

\bibitem{bentley1999three}
P.~Bentley and S.~Kumar, ``Three ways to grow designs: A comparison of
  embryogenies for an evolutionary design problem,'' in {\em Proceedings of the
  1st Annual Conference on Genetic and Evolutionary Computation-Volume 1},
  pp.~35--43, Morgan Kaufmann Publishers Inc., 1999.

\bibitem{dellaert1996developmental}
F.~Dellaert and R.~D. Beer, ``A developmental model for the evolution of
  complete autonomous agents,'' in {\em Proceedings of the fourth international
  conference on simulation of adaptive behavior}, pp.~393--401, MIT Press
  Cambridge, MA, 1996.

\bibitem{eggenberger1997evolving}
P.~Eggenberger, ``Evolving morphologies of simulated 3d organisms based on
  differential gene expression,'' in {\em Proceedings of the fourth european
  conference on Artificial Life}, pp.~205--213, 1997.

\bibitem{hornby2002creating}
G.~S. Hornby and J.~B. Pollack, ``Creating high-level components with a
  generative representation for body-brain evolution,'' {\em Artificial life},
  vol.~8, no.~3, pp.~223--246, 2002.

\bibitem{stanley2009hypercube}
K.~O. Stanley, D.~B. D'Ambrosio, and J.~Gauci, ``A hypercube-based encoding for
  evolving large-scale neural networks,'' {\em Artificial life}, vol.~15,
  no.~2, pp.~185--212, 2009.

\bibitem{stanley2006exploiting}
K.~O. Stanley, ``Exploiting regularity without development,'' in {\em
  Proceedings of the AAAI Fall Symposium on Developmental Systems}, p.~37, AAAI
  Press Menlo Park, CA, 2006.

\bibitem{stanley2007compositional}
K.~O. Stanley, ``Compositional pattern producing networks: A novel abstraction
  of development,'' {\em Genetic programming and evolvable machines}, vol.~8,
  no.~2, pp.~131--162, 2007.

\bibitem{koutnik2010evolving}
J.~Koutnik, F.~Gomez, and J.~Schmidhuber, ``Evolving neural networks in
  compressed weight space,'' in {\em Proceedings of the 12th annual conference
  on Genetic and evolutionary computation}, pp.~619--626, ACM, 2010.

\bibitem{fernando2016convolution}
C.~Fernando, D.~Banarse, M.~Reynolds, F.~Besse, D.~Pfau, M.~Jaderberg,
  M.~Lanctot, and D.~Wierstra, ``Convolution by evolution: Differentiable
  pattern producing networks,'' in {\em Proceedings of the Genetic and
  Evolutionary Computation Conference 2016}, pp.~109--116, ACM, 2016.

\bibitem{schmidhuber1992learning}
J.~Schmidhuber, ``Learning to control fast-weight memories: An alternative to
  dynamic recurrent networks,'' {\em Neural Computation}, vol.~4, no.~1,
  pp.~131--139, 1992.

\bibitem{schmidhuber1993self}
J.~Schmidhuber, ``A ‘self-referential’weight matrix,'' in {\em
  International Conference on Artificial Neural Networks}, pp.~446--450,
  Springer, 1993.

\bibitem{ha2016hypernetworks}
D.~Ha, A.~Dai, and Q.~Le, ``{HyperNetworks},'' in {\em International Conference
  on Learning Representations}, 2017.

\bibitem{younes2010shapes}
L.~Younes, {\em Shapes and diffeomorphisms}, vol.~171.
\newblock Springer Science \& Business Media, 2010.

\end{thebibliography}

\clearpage
% \onecolumn
\appendix

%%%%%%%%%%%%%%%%%%%%%%%%%%%%%%%%%%%%%%%%%%%%%%%%%%%
\section{Reaction-diffusion-advection (RDA) Simulation}\label{sec:rda_sim}

\begin{figure}[!htbp]
\centering
\includegraphics[width=0.8\textwidth]{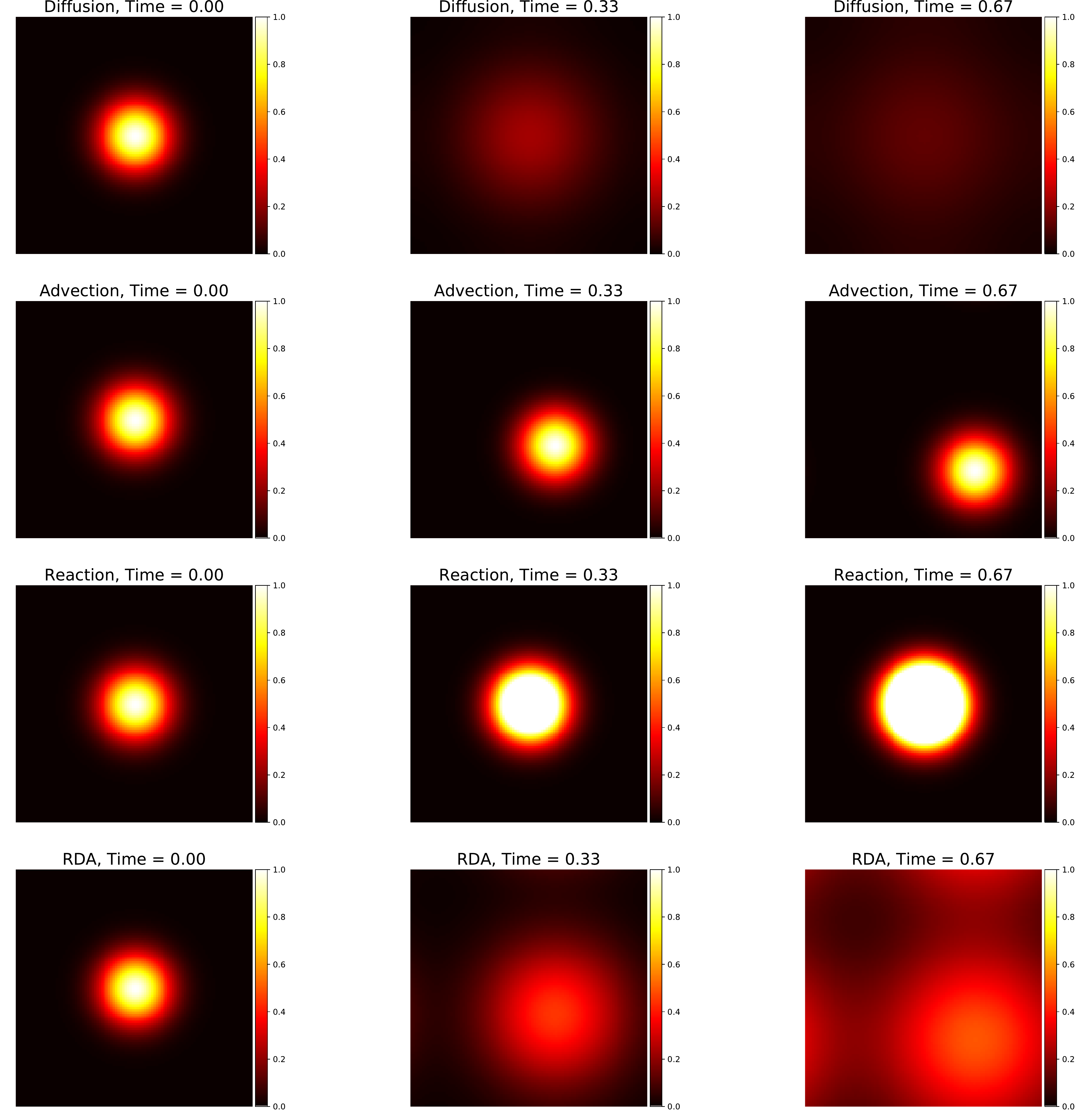}\\
\caption{
Illustration of how different convolution maps could be encoded through
the parameter PDE solver. Here, we show an exemplary convolution at time $t=0$ (left images),
as well as its evolution through time, when we apply the RDA PDE with different settings 
for the model parameters.
Note that with this PDE based encoding, we only need to store the 
initial condition for the parameters (i.e., $t=0$). The rest of the model parameters
could be computed using this initial condition.
}
  \label{fig:RDA_example}
\end{figure}

In this section, we provide details for the reaction-diffusion-advection (RDA) solver as well as an exemplary simulation, shown in~\fref{fig:RDA_example}.
For an illustration of the idea, we set the initial distribution of $\theta$ to be a unit Gaussian centered in the middle of the figure.
In the first row, we show how this single modal Gaussian changes in time when only a diffusion operator is used in the control operator. 
As shown in the first row of the figure, the diffusion operator allows the parameters to evolve from a Gaussian with unit variance to a Gaussian with higher variance.
In the second row, we show how this single modal Gaussian changes with an advection operator. 
Notice how the advection operator allows modeling of different filters centered at different locations with the same variance (since advection operator does not diffuse filters but transports them). 
In the third row, we show how this single modal Gaussian changes when we only use an exponential growth operator for the reaction part. 
Notice how this operator could allow the kernel to increase/decrease its intensities at different pixels in time.
Finally, in the last row, we show a more complex example, where we use all three operators together.

% ------------------------------------------------------------
% ------------------------------------------------------------

%%%%%%%%%%%%%%%%%%%%%%%%%%%%%%%%%%%%%%%%%%%%%%%%%%%%%%%%%%%

\section{Numerical Method}\label{sec:rda_example}

In this section, we provide more details on our numerical techniques.
For simplicity, we set $K$ to be a Dirac delta function, and we use the above RDA function for $q(w,p)$. 
In this case, we have
\begin{equation}\label{eq:final_ode_model}
    \begin{cases}
    \frac{dz}{dt} = f(z; \theta),\\
    \frac{dw}{dt} = \sigma( d \Delta w + \upsilon \cdot \nabla w + \rho w).
    \end{cases}
\end{equation}
Here, we discuss how can we solve the ODE system, given in~\eref{eq:final_ode_model}.
For the evolution of $z$, we follow~\cite{gholami2019anode} and use forward Euler method to solve $z$. For example, if we set time step $N_t$ to be 2, then
\[
z_{1/2} = z_0 + \frac{1}{2}f(z_0; \theta_0);~~~z_{1} = z_{1/2} + \frac{1}{2}f(z_{1/2}; \theta_{1/2}).
\]
It is not hard to see that if $N_t=1$, then the output is the same as the original ResNet.
For the evolution of $\theta$ without non-linearity, i.e., when $\sigma$ is the identity map, there exists an analytic solution in the frequency domain.
Applying the Fast Fourier Transform (FFT) from~\eref{eq:w_rda_pde}, we will get:
\begin{equation}\label{eq:theta_solution}
    F(w)_t =F(d \Delta w + \upsilon \cdot \nabla w + \rho w) ,
\end{equation}
where $F(\cdot)$ denotes FFT operator.
Since the diffusion, advection, and reaction coefficients are constant, we can find the analytical solution in the frequency domain. That is:

\begin{equation}\label{eq:theta_solution_part2}
    w_{t_0+\delta t} = F^{-1}\big( \exp(-\delta t d k^2 + i k\delta t\upsilon +\delta t\rho) F(w_{t_0})\big),
\end{equation}
where $F^{-1}$ is inverse FFT. Note that due to the existence of this analytical solution, the computational cost of solving the evolution for $\theta$ becomes negligible, which is an important benefit of this approach.

When a non-linearity is applied, we use an approximation to solve~\eref{eq:w_rda_pde}, obtaining
\begin{equation}
    w_{t_0+\delta t} =\sigma\left(   F^{-1}\left( \exp(-\delta t d k^2 + i k\delta t\upsilon +\delta t\rho) F(w_{t_0})\right)    \right).
\end{equation}
This means we first apply FFT and its inverse to solve the linear system, and then we apply the non-linear function $\sigma$.
Here, $\delta t$ means the time scale to compute $\theta$.
In this paper, we set the non-linearity function $\sigma$ to be tanh, but other non-linearities could also be used. 

For configuration 1, we use $N_t=5$. 
For configuration 2, we use $N_t = 2$ to solve $z$ and $N_t = 10$ to solve $\theta$. 
In this case, the FLOPS will be only $2\times$ that of the original baseline network. Upon this condition, the process can be formulated as,
\[
z_{1/2} = z_{0} + \frac{1}{2}f(z_0; \theta_0);~~~z_{1} = z_{1/2} + \frac{1}{2}f(z_{1/2}; \theta_1); 
\]
where $\theta_1$ is generated with $\delta t=1/10$.
%%%%%%%%%%%%%%%%%%%%%%%%%%%%%%%%%%%%%%%%%%%%%%%%%%%%%%%%%%%

%%%%%%%%%%%%%%%%%%%%%%%%%%%%%%%%%%%%%%%%%%%%%%%%%%%%%%%%%%%
\section{Model Configuration}\label{sec:model_arch}

In this section, we  provide the architecture we used for the tests in \secref{sec:results}. The AlexNet, ResNet-4, and ResNet-10 we are using are described in following sections.

\subsection{AlexNet}
We used a 2-layer convolution with residual connection added to the second convolution. Thus, we can transform the second convolution into an ODE. Table~\ref{tab:qalexnet_arch} explains detailed structure layer by layer. For simplicity, we omit the batch normalization and ReLU layer added after each convolution. 

\paragraph{Training details} We train AlexNet for 120 epochs with initial learning rate 0.1. The learning rate decays by a factor of 10 at epoch 40, 80 and 100. Data augmentation is implemented. Also, the batch size used for training is 256. Note that the setting is the same for all experiments, i.e., baseline, Neural ODE, and \OURS.

\begin{table*}[!htbp]
\caption{Summary of the architecture used in AlexNet. This is a 2-convolution network with residual connection added to the second convolution, followed by three fully connected layer.}
\small
\label{tab:qalexnet_arch}
\centering
\begin{tabular}{cccccccccccc} \toprule
                      Name  & output size & Channel In / Out  &   Kernel Size & Residual \\
    \midrule
\hc conv1 & 32$\times$32 & 3 / 64 & 5$\times$5 & No \\
\ha max pool & 16$\times$16 & 64 / 64 & - & -  \\
\hc conv2 & 16$\times$16 & 64 / 64 & 5$\times$5 & Yes \\
\ha max pool & 8$\times$8 & 64 / 64 & - & -  \\
\midrule            
  Name &  \multicolumn{2}{c}{input size}  & \multicolumn{2}{c}{output size}  & \\
\midrule
\hc fc1 & \multicolumn{2}{c}{4096} & \multicolumn{2}{c}{384} \\
\ha fc2 & \multicolumn{2}{c}{384} & \multicolumn{2}{c}{192} \\
\hc fc3 & \multicolumn{2}{c}{192} & \multicolumn{2}{c}{10} \\
     \bottomrule 
\end{tabular}
\end{table*}

\subsection{ResNet-4 and ResNet-10}
Here, we provide the architecture of ResNet-4 and ResNet-10 used \secref{sec:results}. We omit the batch normalization and ReLU, for simplicity. Detailed structure is provided in~\tref{tab:resnet}.

\paragraph{Training details} We train ResNet-4/10 for 350 epochs with initial learning rate 0.1. The learning rate decays by a factor of 10 at epoch 150, and 300. Data augmentation is implemented. Also, the batch size used for training is 256. Note that the setting is the same for all experiments, i.e., baseline, Neural ODE, and~\OURS.

\begin{table*}[!htbp]
\caption{Summary of the architecture used in ResNet-4 and ResNet-10. ResNet-10 is a ResNet family that has $2$ layers with $2$ residual blocks in each layer. ResNet-4 has only $1$ layer with only $1$ residual block inside.}
\footnotesize
\label{tab:resnet}
\centering
\begin{tabular}{cccccccccccc} \toprule
                      Name  & output size & Channel In / Out  &   Kernel Size & Residual & Blocks(ResNet-4 / ResNet-10)\\
    \midrule
\hc conv1 & 32$\times$32 & 3 / 16 & 3$\times$3 & No & 1 / 1 \\
\ha \multirow{ 2}{*}{layer1\_1} & \multirow{ 2}{*}{32$\times$32} & \multirow{ 2}{*}{16 / 16} & \multirow{ 2}{*}{\Big[} 3$\times$3 \multirow{ 2}{*}{\Big]} & \multirow{ 2}{*}{Yes} & \multirow{ 2}{*}{1 / 1} \\ &&& 3$\times$3   \\
\ha \multirow{ 2}{*}{layer1\_2} & \multirow{ 2}{*}{32$\times$32} & \multirow{ 2}{*}{16 / 16} & \multirow{ 2}{*}{\Big[} 3$\times$3 \multirow{ 2}{*}{\Big]} & \multirow{ 2}{*}{Yes} & \multirow{ 2}{*}{0 / 1} \\ &&& 3$\times$3   \\
\ha \multirow{ 2}{*}{layer2\_1} & \multirow{ 2}{*}{16$\times$16} & \multirow{ 2}{*}{16 / 32} & \multirow{ 2}{*}{\Big[} 3$\times$3 \multirow{ 2}{*}{\Big]} & \multirow{ 2}{*}{Yes} & \multirow{ 2}{*}{0 / 1} \\ &&& 3$\times$3   \\
\ha \multirow{ 2}{*}{layer2\_2} & \multirow{ 2}{*}{16$\times$16} & \multirow{ 2}{*}{32 / 32} & \multirow{ 2}{*}{\Big[} 3$\times$3 \multirow{ 2}{*}{\Big]} & \multirow{ 2}{*}{Yes} & \multirow{ 2}{*}{0 / 1} \\ &&& 3$\times$3   \\
\midrule            
  Name &  \multicolumn{2}{c}{Kernel Size}  & \multicolumn{2}{c}{Stride}  & Output Size (ResNet-4/ResNet-10)\\
\midrule  
\hc max pool & \multicolumn{2}{c}{8$\times$8} & \multicolumn{2}{c}{8} & 4$\times$4 / 2$\times$2 \\
\midrule            
  Name &  \multicolumn{2}{c}{input size (ResNet-4/ResNet-10)}  & \multicolumn{2}{c}{output size}  & \\
\midrule
\hc fc & \multicolumn{2}{c}{256 / 128} & \multicolumn{2}{c}{10} & \\
     \bottomrule 
\end{tabular}
\end{table*}
%%%%%%%%%%%%%%%%%%%%%%%%%%%%%%%%%%%%%%%%%%%%%%%%%%%%%%%%%%%

%%%%%%%%%%%%%%%%%%%%%%%%%%%%%%%%%%%%%%%%%%%%%%%%%%%%%%%%%%%
\section{Convolution kernel Evolution Example}\label{sec:conv_simulation}

\begin{figure}[!htbp]
\centering
\includegraphics[width=1.\textwidth]{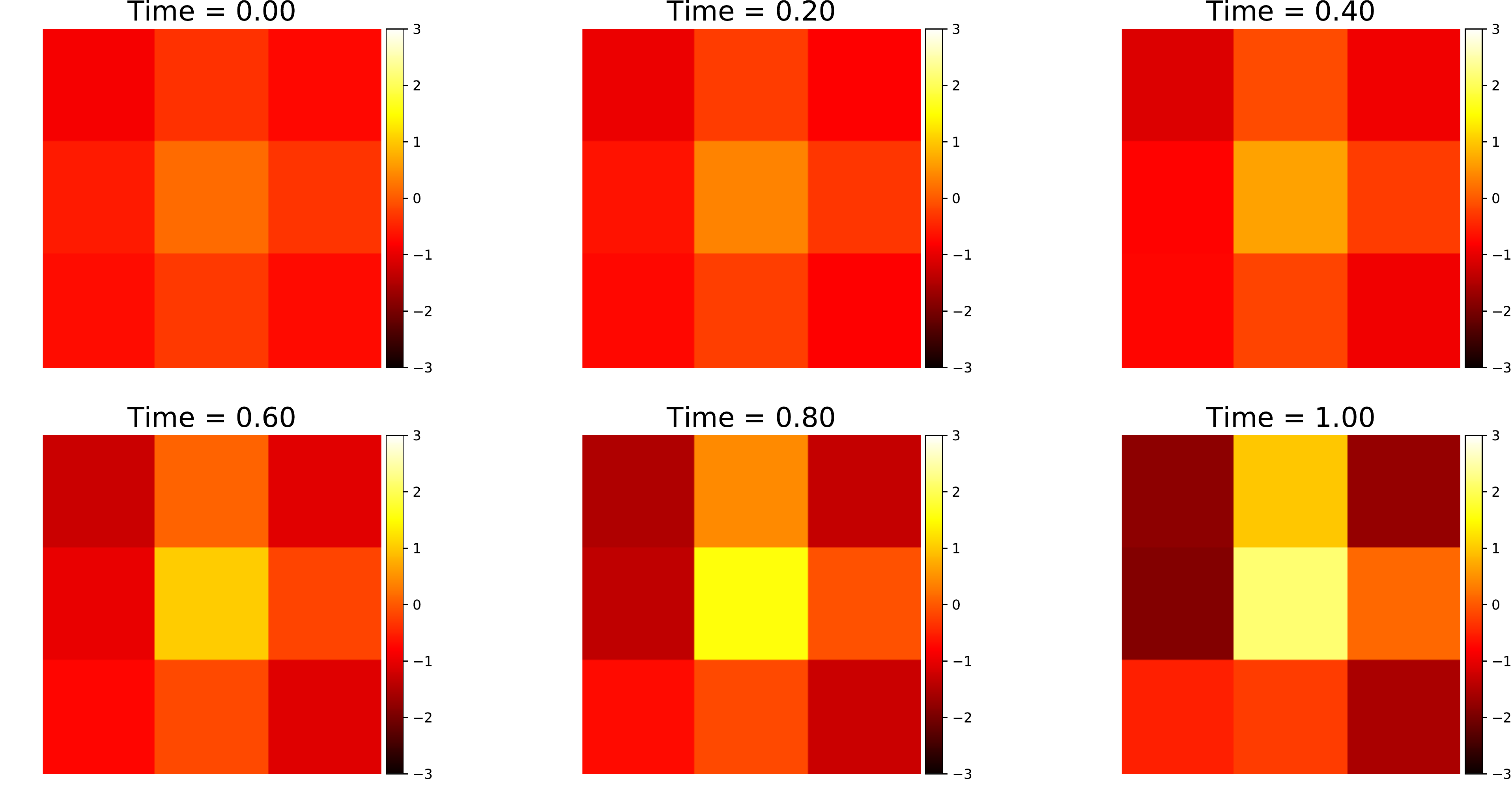}
\caption{
Illustration of how different convolutional operators are evolved in
time during the coupled neural ODE solve (through the evolution operator $q$ in~\eref{eq:anodev2_loss_w_pde}).
The figure corresponds to the first channel of the first convolution kernel parameters of ResNet-4.
These filters will be applied to activation in different time steps (through the $f$ operator in the coupled formulation in~\eref{eq:activation_ode}).
This is schematically shown in~\fref{fig:anode} for three of the filters (the filters are denoted by different shades of brown bars denoted by $\theta$).
Similar pattern can be observed as \fref{fig:qalexnet_kernel}.
}
  \label{fig:resnet4_kernel}
\end{figure}

\begin{figure}[!htbp]
\centering
\includegraphics[width=1.\textwidth]{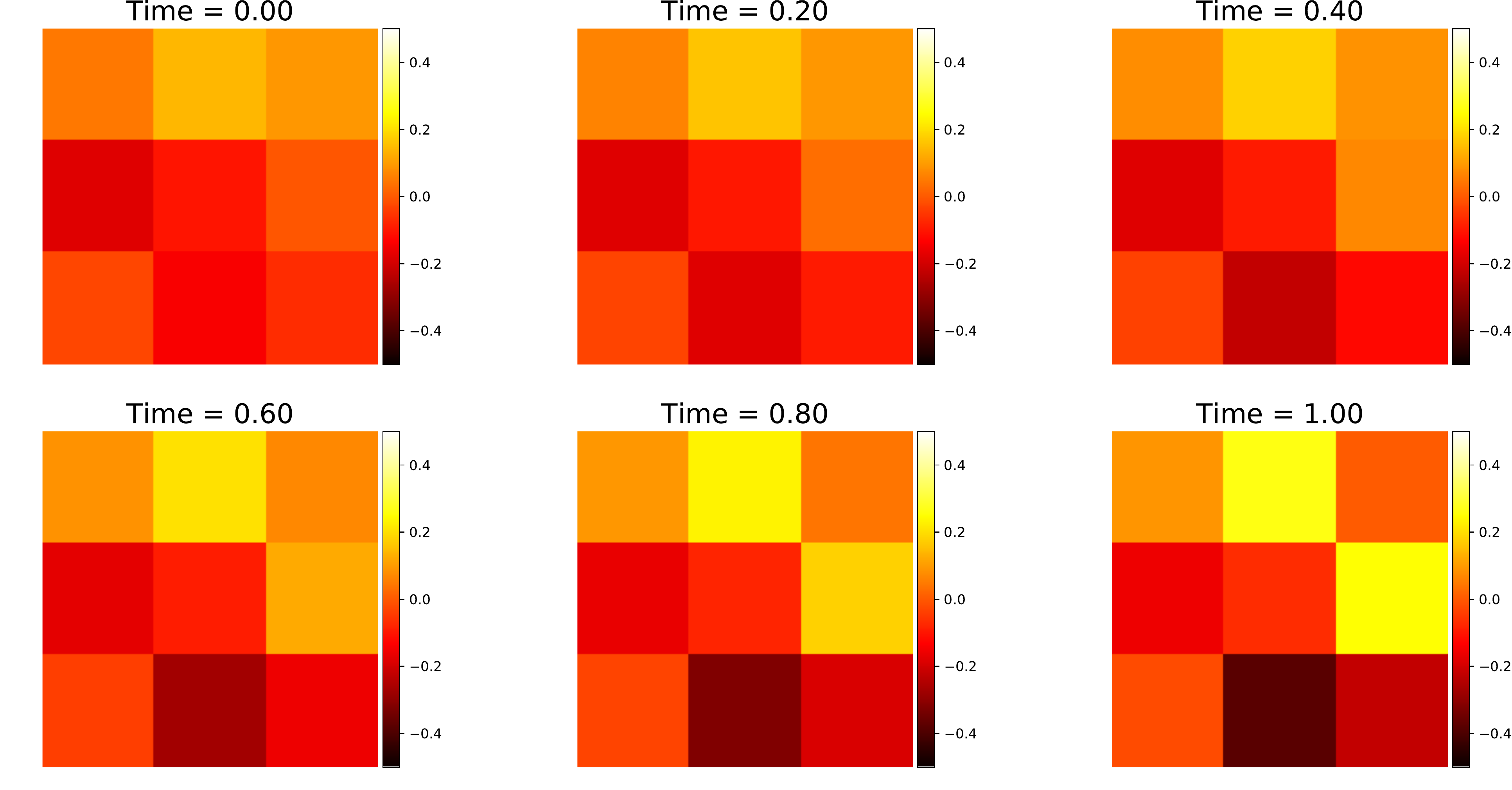}
\caption{
Illustration of how different convolutional operators are evolved in
time during the coupled neural ODE solve (through the evolution operator $q$ in~\eref{eq:anodev2_loss_w_pde}).
The figure corresponds to the first channel of the first convolution kernel parameters of ResNet-10.
These filters will be applied to activation in different time steps (through the $f$ operator in the coupled formulation in~\eref{eq:activation_ode}).
This is schematically shown in~\fref{fig:anode} for three of the filters (the filters are denoted by different shades of brown bars bars denoted by $\theta$).
Similar pattern can be observed as \fref{fig:qalexnet_kernel}.
}
  \label{fig:resnet10_kernel}
\end{figure}

In this section, we show some examples of how the model parameters $\theta$ are evolved in time. Results for ResNet-4 and ResNet-10 are shown in~\fref{fig:resnet4_kernel} and~\fref{fig:resnet10_kernel}, respectively. 

%%%%%%%%%%%%%%%%%%%%%%%%%%%%%%%%%%%%%%%%%%%%%%%%%%%%%%%%%%%

\section{Derivation of Optimality Conditions}\label{sec:opt_condtion}

In this section, we present a detailed derivation of the optimality conditions corresponding to~\eref{eq:anodev2_loss}.
We need to find the so-called KKT conditions, which can be found by finding stationary points of the corresponding Lagrangian, defined as:
\begin{align}
\begin{split}
    \gL &= \gJ(z_1) + \int_0^1\alpha(t) \cdot \left(\frac{dz}{dt} - f(z(t),\theta(t))\right)dt 
    +\int_0^1\beta(t)\cdot \left(\frac{\partial w}{\partial t} - q(w, p)\right)dt \\
    &+\int_0^1\gamma(t) \cdot \left(\theta(t) - \int_0^t K(t-\tau)w(\tau)d\tau\right)dt
    % + \alpha_0\cdot(z_0 - z(0)) + \beta_0\cdot(w_0 - w(0)).
\end{split}
\end{align}

In order to derive the optimality conditions, we first  take variations with respect to $\alpha(t)$, $\beta(t)$, and $\gamma(t)$. 
This basically results in the ``Activation ODE'', the ``Evolution ODE'', and the relation between $\theta(t)$ and $w(t)$, shown in~\eref{eq:anodev2_loss}.
Taking variations with respect to $z(t)$ will result in a backward-in-time ODE for the $\alpha(t)$, which is continuous equivalent to backpropagation (i.e., Optimize-Then-Discretize).
Taking variations with respect to $\theta$ will result in an algebraic relation between $\alpha(t)$ and $\gamma(t)$.
Taking variations with respect to $w(t)$ will be split in two parts: variations with respect to $w(t)$ for $t>0$; and variations with respect to $w(0)$, which is in fact one of our unknown parameters.
The split is performed by first integrating by parts the $\int_{0}^1 \beta(t) \frac{\partial w(t)}{\partial t}dt$
term to expose a term that reads $\beta(1)w(1)-\beta(0)w_0$,
and then taking variations with respect to $w_0$.
Finally, we also need to take variations with respect to the vector $p$.
An important technical detail is that to take the variations of the $\int_0^1 \gamma(t) \cdot \int_{0}^t K(t-\tau)w(\tau)d\tau dt $
with respect to $w$ can be done easily by converting it to $\int_{0}^1 \gamma(t)\cdot \int_0^1 H(\tau - t)K(t-\tau)w(\tau)d\tau dt$.
The details are given below.

%%%%%%%%%%%%----------------------%%%%%%%%%%%%%%%%%
In order to satisfy the first optimality condition on $z$ we have:
\[
\left(\frac{\partial \gL}{\partial z}\right)^T\hat z =0,
\]
where this equality must hold for any variation $\hat z$ in space and time.
We have:
\begin{align}
\begin{split}
    \left(\frac{\partial \gL}{\partial z}\right)^T\hat z 
    &= \left(\frac{\partial \gJ(z_1)}{\partial z_1}\right)^T \hat z_1 + \alpha_1^T\hat z_1 + \int_0^1\left(-\frac{\partial \alpha}{\partial t} - \frac{\partial f(z,\theta)}{\partial z}^T\alpha\right)^T\hat z dt  = 0.
\end{split}
\end{align}
Imposing this condition holds for all variation $\hat z$ will result in the first adjoint equation as follows:
\begin{equation}
    \frac{\partial \gJ(z_1)}{\partial z_1} + \alpha_1=0, ~~~~ -\frac{\partial \alpha}{\partial t} - \Big(\frac{\partial f}{\partial z}\Big)^T\alpha = 0.
\end{equation}
%%%%%%%%%%%%----------------------%%%%%%%%%%%%%%%%%

%%%%%%%%%%%%----------------------%%%%%%%%%%%%%%%%%

For $\theta$, the following equation needs to be satisfied:
\[
\left(\frac{\partial \gL}{\partial \theta}\right)^T\hat \theta =0.
\]
We have
\begin{equation}
    \left(\frac{\partial \gL}{\partial \theta}\right)^T\hat \theta= \int_0^1\left(-\frac{\partial f(z,\theta)}{\partial \theta}\right)^T\alpha^T \hat\theta dt +\int_0^1\gamma^T\hat\theta dt.
\end{equation}
This further implies:
\begin{equation}
    -\frac{\partial f(z,\theta)}{\partial \theta}^T\alpha + \gamma = 0.
\end{equation}
%%%%%%%%%%%%----------------------%%%%%%%%%%%%%%%%%

%%%%%%%%%%%%----------------------%%%%%%%%%%%%%%%%%
Finally, the inversion equation on $w$ could be found by performing variation on $w$:
\[
\left(\frac{\partial \gL}{\partial w}\right)^T\hat w =0.
\]
We have 
\begin{align}
\begin{split}
    \left(\frac{\partial \gL}{\partial w}\right)^T\hat w &=  \int_0^1-\left(\frac{\partial \beta}{\partial t} - \frac{\partial q(w; p)}{\partial w}^T\beta\right)^T\hat w dt \\
    &+  \beta_1^T\hat w_1 +\int_{0}^1  \int_0^t- H(\tau - t)K^T(t-\tau)\gamma d\tau \hat w dt \\
    &= \int_0^1-\left(\frac{\partial \beta}{\partial t} - \frac{\partial q(w; p)}{\partial w}^T\beta\right)^T\hat w dt \\
    &+  \beta_1^T\hat w_1 +\int_{0}^1\int_0^t-H(\tau -t)K^T(t-\tau)\gamma d\tau \hat wdt,
\end{split}
\end{align}
where $H$ is the scalar Heaviside function.
Imposing that this condition holds for all variation $\hat w$ will result in the inversion equation as follows,
\begin{equation}
     -\frac{\partial \beta}{\partial t} - \frac{\partial q(w; p)}{\partial w}^T\beta + \int_t^1- K^T(t-\tau)\gamma d\tau,~~~\beta_1 = 0.
\end{equation}
%%%%%%%%%%%%----------------------%%%%%%%%%%%%%%%%%

The gradient of $\gL$ with respect to $w_0$ can be computed as,
\begin{align}
\begin{split}
    g_{w_0} = \frac{\partial \gL}{\partial w_0} =  \frac{\partial R(w_0, p)}{\partial w_0} - \beta_0.
\end{split}
\end{align}

Finally, the gradient of $\gL$ with respect to $p$ can be computed as,
\begin{align}
\begin{split}
    g_p = \frac{\partial \gL}{\partial p} = \frac{\partial R(w_0, p)}{\partial p} - \int_0^1\frac{\partial q(w,p)}{\partial p}^T \beta(t) dt.
\end{split}
\end{align}

Note that if optimality conditions are achieved with respect to $w_0$ and $p$, then 
\begin{equation}
    g_{w_0} = 0,~~~g_p = 0.
\end{equation}
% %%%%%%%%%%%%----------------------%%%%%%%%%%%%%%%%%

\end{document}